\documentclass[letterpaper, 10pt, conference]{ieeeconf}  

\IEEEoverridecommandlockouts                              

\usepackage{multirow}
\usepackage{amsmath}
\usepackage{graphicx}
\usepackage{tabularx}
\usepackage[table]{xcolor}
\newcolumntype{Y}{>{\centering\arraybackslash}X} 
\usepackage{boldline}
\usepackage{paralist}
\usepackage{tabularx}
\usepackage{colortbl}
\usepackage{caption}
\usepackage{algorithm}
\usepackage{microtype}
\usepackage[noend]{algorithmic}
\usepackage{hyperref}
\newcommand{\para}[1]{\noindent\textbf{#1.}}
\newcommand{\figurename}{Figure}
\newcommand{\tablename}{Table}
\usepackage{filecontents}
\usepackage[noadjust]{cite}

\title{\LARGE \bf
HGIC: A Hand Gesture Based Interactive Control System for Efficient and Scalable Multi-UAV Operations
}

\author{Mengsha Hu$^{1, 2}$, Jinzhou Li$^{1, 3}$, Runxiang Jin$^{1}$, Chao Shi$^{4}$, Lei Xu$^{2}$, Rui Liu$^{1*}$
\thanks{$^{1}$ Cognitive Robotics and AI Lab (CRAI), College of Aeronautics and Engineering, Kent State University, Kent, OH 44240, USA. $^{2}$ Department of Computer Science, Kent State University, Kent, OH 44240, USA. $^{3}$ College of Engineering, Cornell University, Ithaca, NY 14853, USA. $^{4}$ Department of Systems Science and Industrial Engineering, Binghamton University, Binghamton, NY 13902, USA.}
\thanks{$^{*}$ Corresponding author, email: {\tt\small ruiliu.robotics@gmail.com}.
}
}

\begin{document}

\maketitle
\thispagestyle{empty}
\pagestyle{empty}

\begin{abstract}

As technological advancements continue to expand the capabilities of multi unmanned-aerial-vehicle systems (mUAV), human operators face challenges in scalability and efficiency due to the complex cognitive load and operations associated with motion adjustments and team coordination. Such cognitive demands limit the feasible size of mUAV teams and necessitate extensive operator training, impeding broader adoption. This paper developed a Hand Gesture Based Interactive Control (HGIC), a novel interface system that utilize computer vision techniques to intuitively translate hand gestures into modular commands for robot teaming. Through learning control models, these commands enable efficient and scalable mUAV motion control and adjustments. HGIC eliminates the need for specialized hardware and offers two key benefits: 1) Minimal training requirements through natural gestures; and 2) Enhanced scalability and efficiency via adaptable commands. By reducing the cognitive burden on operators, HGIC opens the door for more effective large-scale mUAV applications in complex, dynamic, and uncertain scenarios. HGIC will be open-sourced after the paper being published online for the research community, aiming to drive forward innovations in human-mUAV interactions.

\end{abstract}


\section{Introduction}
With advances in manufacturing, sensors, control, and artificial intelligence algorithms, a multi-UAV system (mUAV), where multiple UAVs provide diverse functionalities such as perceiving, maneuvering, delivering, and communication support,
has been implemented in wide-range applications, from military operations and urban security to disaster response \cite{9214446,8695011, 8715365}. However, the operation of mUAV is cognitively intensive, as a human operator needs to adjust motions for individual robots while maintaining collaborative strategies for the team in tasks like chasing targets and searching areas. Considering a human has limited cognitive bandwidth, the affordable, controllable size of mUAV teams will be limited, which constrains the efficiency, scale, and usage of mUAV in real applications \cite{Sargolzaei2020, 4445761, 9900763}. Moreover, due to the coordination nature, operating an mUAV is complex; a human operator needs to receive intensive prior training to operate a mUAV, which will further discourage mUAV usage \cite{HRIreview, 10.1145/2750858.2805823, gao2012teamwork}. Therefore, these combined challenges highlight the urgent need for a solution that can intuitively interpret human commands into mUAV precise mission executions, breaking through the limitations of slow and complex human command delivery and multi-robot control.

\begin{figure}
    \centering
    \includegraphics[width=0.95\linewidth]{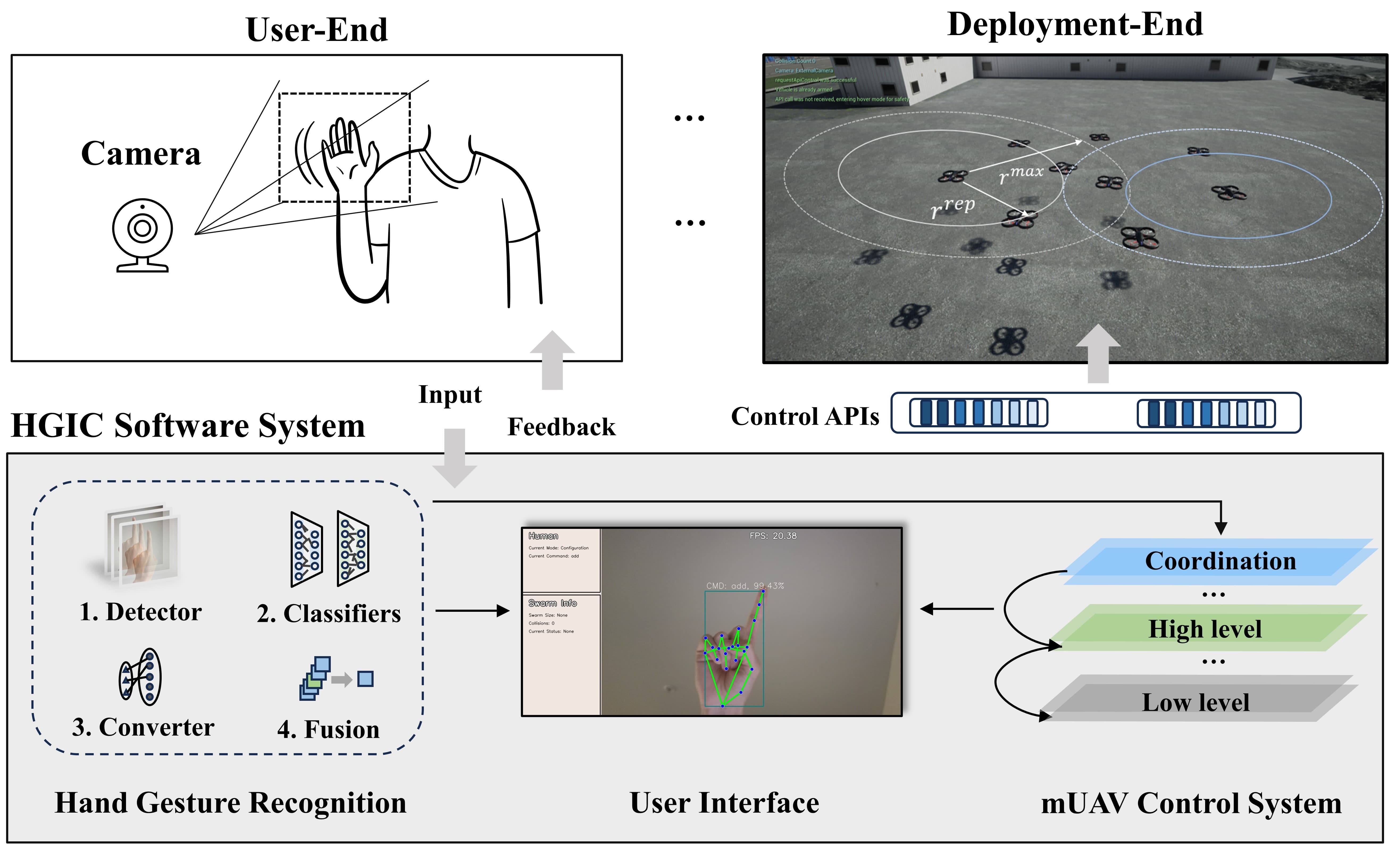}
    \caption{Architecture illustration for hand gesture interactive control (HGIC) system.}
    \label{fig:overview}
\end{figure}

In our work, we introduce a novel system: Hand Gesture Based Interactive Control (HGIC) designed to operate Multi-Robot Systems (MRS) efficiently and scalably. An overview of the framework is depicted in \figurename~\ref{fig:overview}. At the core of HGIC lies an integration of camera-based gesture recognition, an intuitive command library, a multi-layered distributed control architecture, and an interactive user-feedback interface. Through computer vision technologies, human commands will be naturally expressed by hand gestures and translated into modular commands, so operators can easily adjust team sizes or change tasks for the MRS. With learning control models, such as meta-learning control \cite{huang2021meta}and reinforcement learning control \cite{9406813}, commands will be translated to teaming adjustments for an MRS. This design 1) eliminates the dependency on specialized hardware, 2) reduces training time for pilots, and 3) ensures efficient coordination among UAVs. Designed for surveillance, search, and target-chasing tasks, HGIC optimizes mUAV operations, ensuring its performance while minimizing the cognitive load on the human operator. 

Our contributions to this paper include:
\begin{itemize}

\item A gesture recognition system enables intuitive control of mUAVs with natural body language without specialized control hardware.

\item A mode-based gesture library organized by five intuitive modes - navigation, formation, task, configuration, and safety - supporting efficient execution of diverse complex behaviors with limited gesture sets.

\item A user interface for real-time interactions between humans and mUAVs.

\item A multi-layer distributed mUAV control architecture that interprets high-level user commands and autonomously coordinates intricate teaming strategies and collision avoidance, enabling operators to focus on high-level mission goals.

\item APIs for flexible mUAV configuration and coordination, allowing seamless modification and deployment.

\end{itemize}

\section{Related Work}
\para{Interaction Modalities} Nowadays, powered by AI technologies such as video understanding and speech recognition, there have been significant advances in human-robot interaction methods. These encompass modalities including: 1) hand/Arm/body gestures-based interface, where camera devices were used to understand human instructions across a wide range, such as desired trajectories and specific actions \cite{gesture, gesture2, HandMotion, body}; 2) Touch-based interface, where tactile sensors \cite{SwarmTouch, touch2}, or joysticks \cite{4797870, 1389777} were utilized to retrieve human feedback, such as desired destinations and movement directions; 3) Voice-based, where audio speech was received by microphones and processed by natural language understanding and AI techniques to extract command information \cite{voice1, voice2}; 4) Eye-based signals, where camera devices and AI techniques were applied to track eyeball movements and physical status (e.g., eye open/close/blinking) to interpret human indications \cite{9372809, yu2014human}; 5) Multi-modality, where multiple-type sensors (e.g.,visual-audio sensors) were used to interpret human commands \cite{9981165, maza2010multimodal}; and 6) Virtual/augmented reality interface \cite{9089444, chen2021pinpointfly}, where VR/AR glasses were used to create immersive mission environments for a better sense of space and motion for remote robot/UAV control. Each method effectively delivers human instructions to robots, each with unique advantages.

\textit{Physical Signal Methods (Body Motion, Eyeball Movements, Tactile).}
Advantages: These methods naturally interact with humans in a way that is consistent with daily social communications, thus reducing the need for prior training. They are also cost-effective, relying on simple camera devices.
Disadvantages: Limited by their symbolic simplicity, these methods can only convey short and basic commands. Unlike broader body language or arm movements, hand gestures can convey a richer range of meanings due to the flexible capabilities of human hands and fingers.

\textit{Speech Methods.}
Advantages: These methods enable natural human-robot interaction and, thanks to the richness of human speech, can express a wide scope of complex commands. This is particularly useful for tasks like robotic furniture assembly or construction.
Disadvantages: Speech ambiguity makes it difficult for robots to understand human commands without accurate contextual cues accurately. Although not as information-rich as speech, gestures can provide clearer spatial and directional information, reducing unnecessary back-and-forths and preserving human cognitive bandwidth.

\textit{VR/AR Methods.}
Advantages: These methods offer more precise commands because they use accurate 3D maps and landmarks.
Disadvantages: The requirement for specialized equipment, like expensive glasses and complex map designs, limits their broader application.

Many of the existing interaction techniques are primarily as conduits for human commands without assisting in robotic decision-making, thus adding to the human cognitive load. HGIC is integrated with various mUAV planning algorithms, adjusts mUAV coordination strategies based on human input, and is adaptable to a wide scope and scale of mUAV teams.

\para{Multi-Robot Control} 
The Reynolds flocking algorithm \cite{reynolds1987flocks} is one of the early distributed control algorithms that realized simple collision avoidance and flocking of UAVs. Later, approaches to handle more complex tasks and formations were proposed. For instance, a behavior-based method was provided in~\cite{teambehavior}. The recent work in~\cite{vasarhelyi2018optimized} showed stable swarm behavior in simulations and was confirmed with 30 drones in field tests. In terms of control system architecture design, the early work in \cite{layer1} adopted layered architectures for hierarchical control, but it had some limitations in scalability. Recent work proposed a multi-level distributed fusion architecture based on missions \cite{9265199}, which considered layered control and system scalability simultaneously. Drawing inspiration from \cite{9265199}, we propose a simplified multi-level distributed control architecture combining biologically inspired swarm behavior \cite{reynolds1987flocks,teambehavior, schilling2018learning}. Furthermore, the new architecture also considers the autonomy of lower-level coordination control to ensure the safety and robustness of the system.

\section{Module Architecture}
In this section, we present the system architecture of the human-gesture Interaction Control(HGIC), which integrates a hand gesture recognition system with a mUAV control system. The major components of the HGI platform and their relationships are summarized in \figurename~\ref{fig:overview}.

\para{Module I: Hand Gesture Recognition system (HGR)}
HGR identifies the operator's hand gestures. Upon recognizing specific gestures, the system interprets the results into corresponding control commands for the mUAV. 

\para{Module II: mUAV Control System}
The Control System is an autonomous distributed system centered around a mUAV controller that operates based on distributed algorithms. It is equipped with built-in high-level tasks and formation configurations and has the ability to automatically adjust control in response to real-world conditions, including tasks such as path planning and obstacle avoidance.

\begin{figure}[!t]
    \centering
    \includegraphics[width=\linewidth]{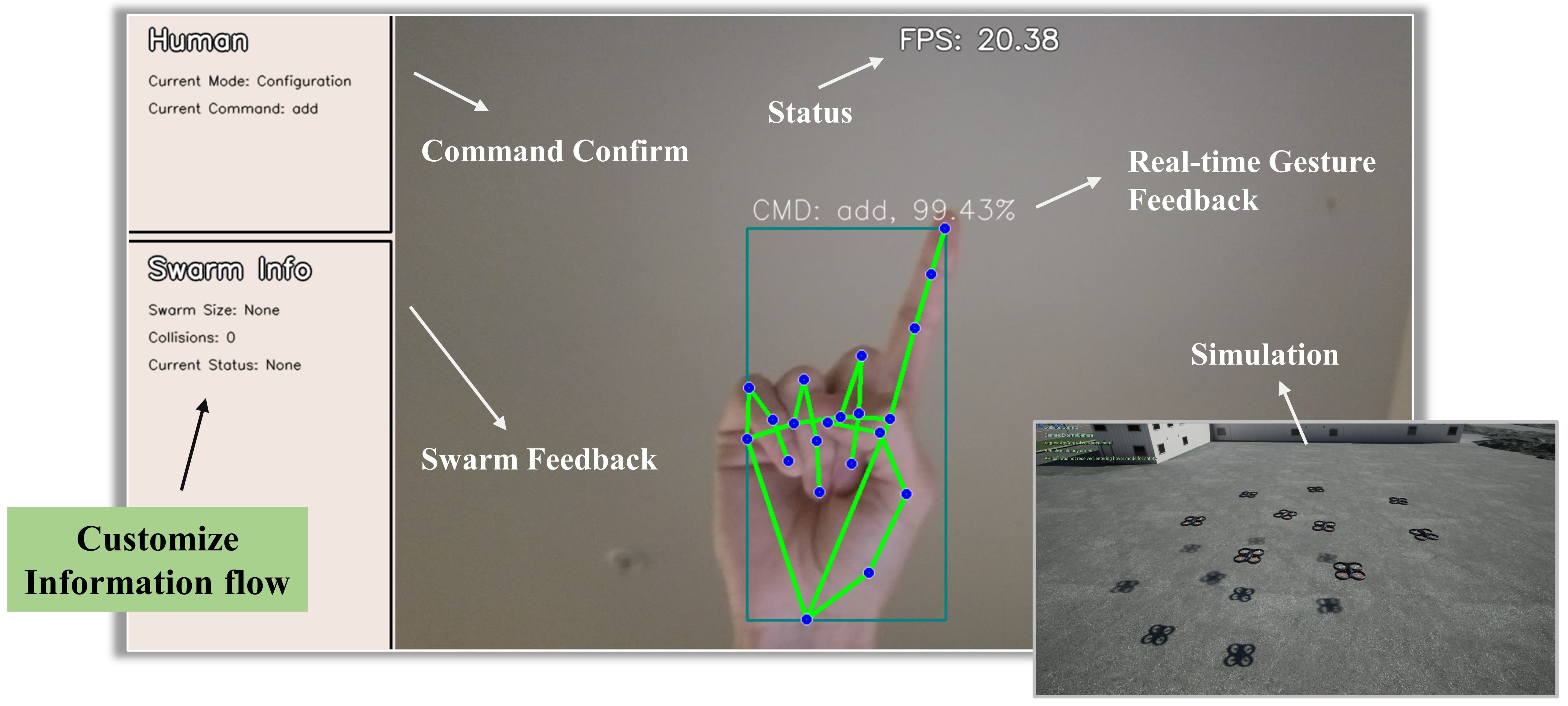}
    \caption{HGIC graphical user interface. The GUI is divided into three main sections: 1) Current command feedback; 2) mUAV feedback; and 3) Real-time feedback for gesture recognition. Additionally, the frame rate is displayed at the top side.} 
    \label{fig:ui}
\end{figure}

\para{Module III: User-Interface}
Our user interface, as shown in \figurename~\ref{fig:ui}, is designed with a focus on three key aspects: \textbf{a).} The current actions of the human operator, including engaged commands and modes; \textbf{b).} The feedback from mUAV, such as size, formation, and collision count; and \textbf{c).} The real-time feedback of user gesture recognition allows users to double-check their gestures. To balance the information flow of operation, we integrate the APIs from the AirSim simulation environment, enabling users to customize the display of information flows with minor modifications. This design encourages users to explore and compare various information combinations, like collision locations, collision counts, speed, and UAV group size, to determine their best display strategy.

\para{Module IV: Communication Support} 
We implement a two-way communication mechanism using the UDP-Socket protocol in our simulation to enable seamless communication between the mUAV team and the interaction system. Human operators can convey control commands through intuitive gestures, directly influencing the actions of the mUAV. Simultaneously, the mUAV feeds back real-time execution status, creating a dynamic loop of information.

\para{System Design Strategy} 
Within the broad design process of the system, we emphasize the principles of modular structure and scalability. Adopting a modular approach ensures that each critical component is independently developed and tested, thereby promoting efficiency and adaptability. Our system design simplifies not only future upgrades but also customization to meet specific needs. Furthermore, we develop a flexible converter which is essentially a set of mapping rules that can easily accommodate various interaction modalities with only minor adjustments. This design ensures that the system can be easily integrated with emerging technologies such as Virtual Reality (VR) and Augmented Reality (AR) and also ensures that any future advancements remain compatible and up-to-date. 

\para{Deployment} We encapsulate the control algorithms within the interface, ensuring users can execute complex behaviors by invoking our APIs, such as high-level coordination strategies (e.g., area coverages, formations, flocking, slitting/merging).



\begin{figure*}[!t]
    \centering
    {\includegraphics[width=\textwidth]{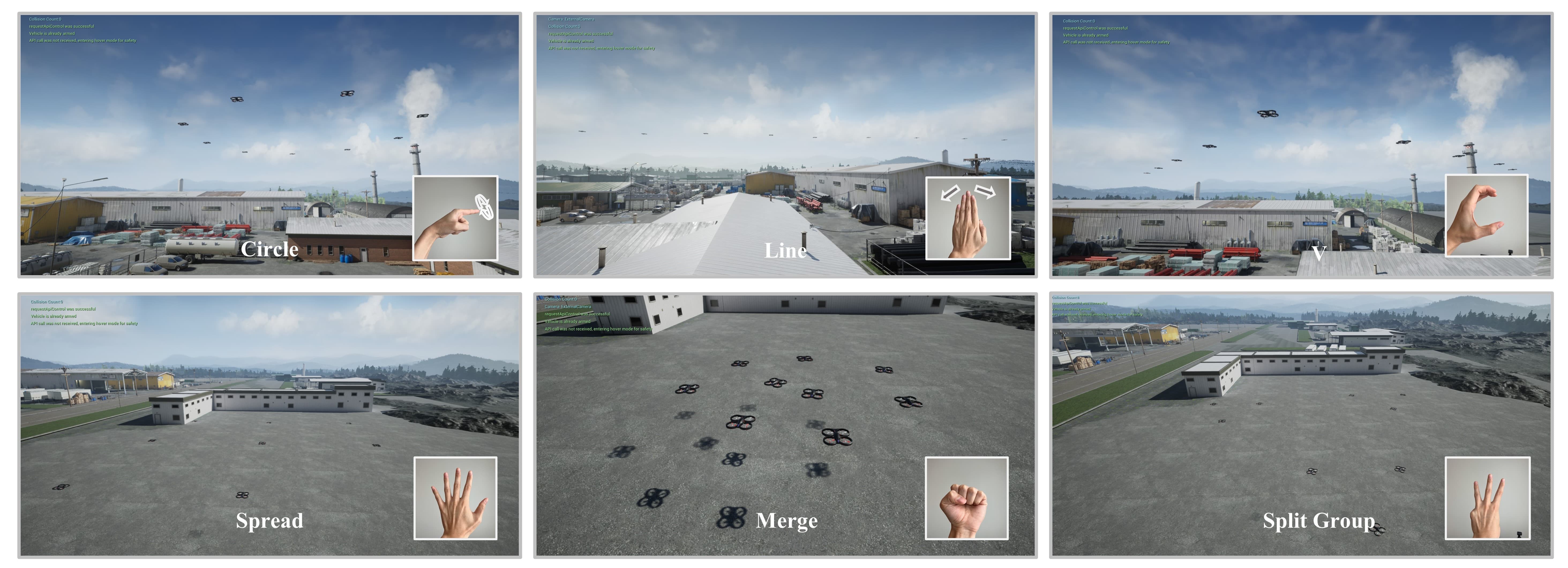}}
    \centering
    {\includegraphics[width=\textwidth]{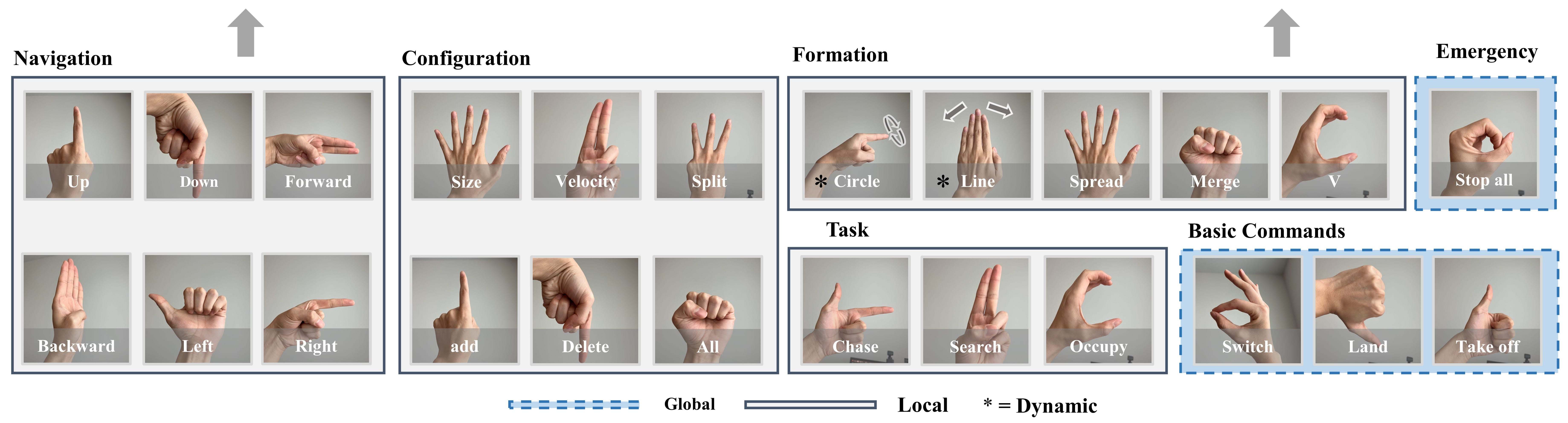}}
    \caption{Gestures for swarm control. The upper part of the figure shows various formations/tasks adopted by the swarm, including evolving from circles to lines, culminating in a V-shape, expanding, merging, and splitting into three distinct groupings. 
    The bottom part summarizes the common gestures for multi-UAVs control.
    }
    \label{fig:gesture-swarm-control}
\end{figure*}

\section{Human Gesture Command Interpreting Module}
The Human-Gesture Recognition (HGR) system is designed to support real-time interaction between human operators and multi-UAVs. The hardware of HGR comprises an RGB camera and a computer equipped with the Nvidia GeForce RTX 3080. The RGB camera captures the user's gestures while the computer processes the captured image data and operates the AirSim~\cite{airsim2017fsr} simulation of the swarm behavior. Notably, a smartphone can easily replace the RGB camera, which greatly improves the system's accessibility and portability.

\subsection{Hand Gesture Detector}
The detector utilizes image frames captured by the RGB camera as input. We use MediaPipe hand tracking pipeline~\cite{lugaresi2019mediapipe,zhang2020mediapipe} to predict the hand skeleton of a human, all without the need for any specialized equipment. Firstly, the detector scans the image to locate the hand within the frame. This involves pinpointing the hand's position and orientation and encapsulating it within a bounding box. Once the hand's position is identified, the detector extracts 21 hand key points. These points correspond to anatomical features of the hand, such as the fingertips and the center of the palm. They are represented as $(x, y)$ coordinate pairs in 2D space, clearly describing the hand structure.

\subsection{Hand Gesture Classifier}
Once the coordinate data of the gestures are obtained, the Classifier is responsible for interpreting and recognizing these gestures. To better distinguish static and dynamic gestures, we build two neural network classifiers~\cite{sung2021device}, one specializing in spatial patterns and the other in temporal patterns. Both classifiers rely on hand-skeletal data~\cite{Smedt_2016_CVPR_Workshops, 8373818, lugaresi2019mediapipe} as input and exclude the consideration of depth images. The Static Classifier is designed to process the hand's spatial configuration at discrete moments. The Dynamic Classifier leverages the time series information of the gestures. It focuses on how hand elements change over time throughout the entire gesture.

\subsection{Hand Gesture Command Converter} 
The command converter lies in a predefined library that translates human instructions into executable directives for UAVs. To create a flexible interface between the user and the controller, we provide a modifiable file, formatted in JSON, that enables users to modify and define the mapping rules, eliminating the need for extensive modification. This design also ensured the interaction's extensibility, laying the foundation for future integration with forthcoming Virtual Reality (VR) and Augmented Reality (AR) approaches. 

To improve the efficiency, flexibility, and safety of controlling mUAV, we integrate a hierarchical, modular, and user-centered approach to design our gesture library. We propose a command library with five modes and two types of priorities to support the operator to use limited gestures for complex tasks and the large number of UAVs. 
The five modes are:
\begin{inparaenum}
    \item Navigation Mode for basic motion control;  
    \item Formation Mode to organize predefined formations;  
    \item Task Mode to execute complex tasks such as area search, space occupation and target chasing; 
    \item Configuration Mode to adjust UAV size and operation parameters, enabling the selection of appropriate mUAV for specific tasks; and 
    \item Safety Mode for emergency control requirements. 
\end{inparaenum}
And the two types of command with different priorities are:
\begin{inparaenum}
    \item Local Commands are restricted to individual modes for precise operations, which have lower priority; and
    \item Global Commands operate across modes to support high-level tasks, which have higher priority and are especially useful in emergency situations.
\end{inparaenum}
\figurename~\ref{fig:gesture-swarm-control} illustrates the entire gesture commands and modes, offering an overview of the functionality in our system.

\subsection{Hand Decision Fusion}
Due to the fast processing capability of gesture recognition, a substantial amount of command information can quickly be fed into the system, which brings challenges for following command interpretation and UAV swarm control. To address this problem, we designed a command buffering mechanism. This mechanism only permits gestures with an ideal recognition rate $(P)$ to be added to the buffering queue $(N)$. We then utilize a time-weighted strategy for each gesture within the buffer and aggregate the weights of identical gestures to get the final decision. This allows the system to understand better and respond to the user's most recent intentions. Moreover, to prevent sending commands repeatedly, we set a frame blank period $(F)$ between two consecutive commands. In our implementation, we chose $P\geq=0.9$, $N=20$, and $F=50$.

\section{mUAV Learning Control Module and APIS}

\begin{figure}[t]
    \centering
    \includegraphics[width=\linewidth]{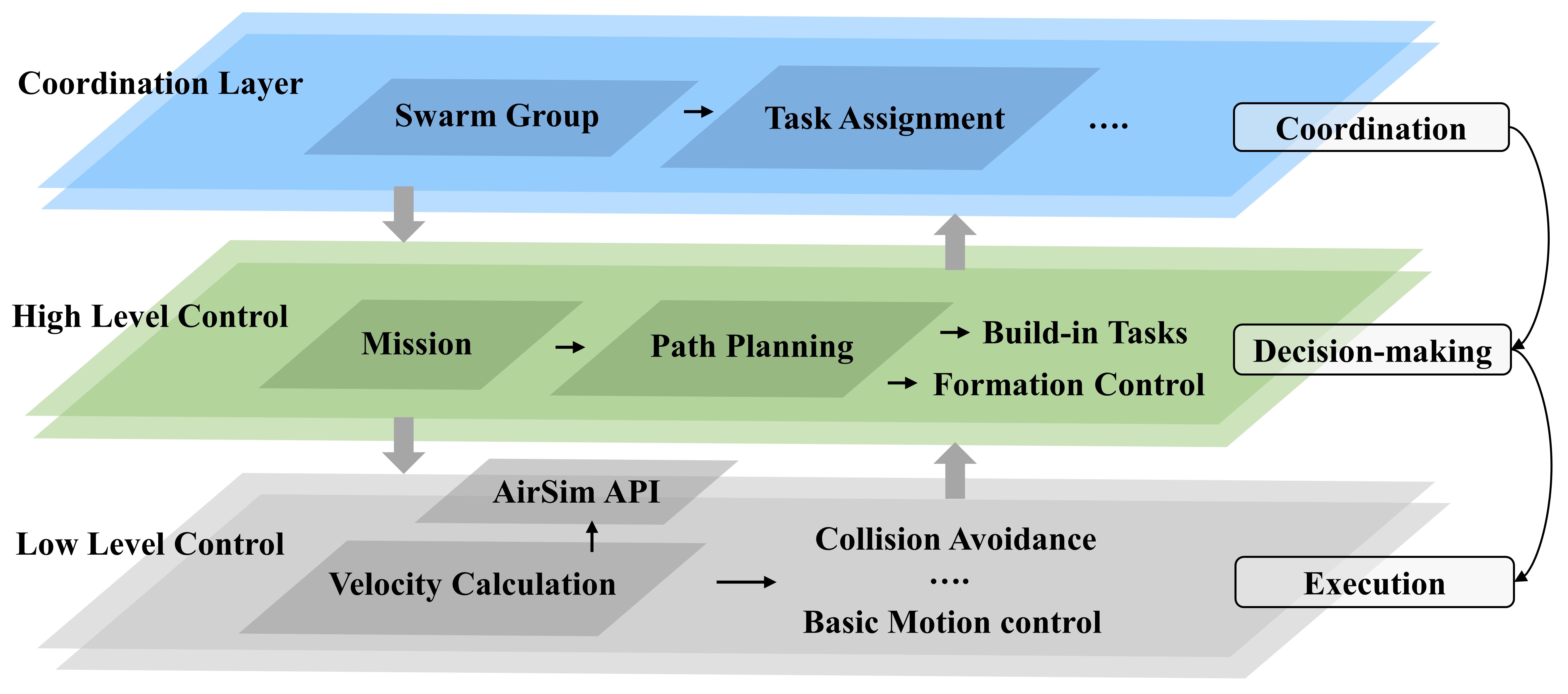}
    \caption{mUAV system control architecture. The architecture comprises of three-levels, basic motion control at the low level, advanced task and formation management at the high level, and task allocation at the coordinated control level.}
    \label{fig:control}    
\end{figure}

\subsection{Multi-Level Distributed Control Architecture}
The UAV swarm control system is designed with a hierarchical, distributed architecture composed of three main layers: Low-Level Control, High-Level Control, and Coordination, which is inspired by~\cite{9265199} and depicted in \figurename~\ref{fig:control}. 
This structure ensures coordination while maintaining flexibility, scalability, and fault tolerance. 

\para{Low-Level Control} 
This layer is responsible for managing basic flight dynamics including stabilization, speed control, collision avoidance, and other reflexive behaviors critical to maintaining safe flight. It interprets abstract goals received from higher layers and executes these commands by calculating the velocities. 
This layer is designed to focus only on immediate control and does not require knowledge of the broader mission planning or strategy. To achieve this, we leveraged the AirSim API~\cite{airsim2017fsr}, which provides a standardized API for simulating dynamics, actuator commands, and onboard sensor inputs without the need for detailed control implementation. 
For example, it processes sensor data, calculates repulsive forces, separation forces, cohesion forces, migration velocities, etc., and converts these forces into velocity commands.

\para{High-Level Control} 
This layer manages various predefined advanced tasks and different UAV formations to adapt to diverse environments while concurrently handling swarm algorithm logic. Upon receiving task instructions from the Coordination layer, the high-level control calculates each UAV's relative position and speed parameters in the selected formation, determining the spatial deployment of multiple UAVs. The results are then transmitted to the low-level execution via a standard interface. Simultaneously, the high-level control continuously monitors the low-level execution and adjusts the task execution plan as necessary to ensure task requirements are fulfilled. This design enables the high-level control to efficiently coordinate all UAVs in the preset type library, manage and coordinate various complex tasks under human command, and real-time monitor and adjust the low-level execution, ensuring successful task completion.

\para{Coordinated Control} 
This layer is responsible for UAV grouping and task allocation based on human commands. This is a pivotal layer as it directly impacts the efficiency of the entire UAV swarm. The operator can dynamically allocate tasks and groups according to the current situation and objectives. 

\subsection{Advanced Task-Planning Functions}
Advanced tasks involve a series of built-in complex tasks that are fundamental to achieving mission objectives in various scenarios. These tasks are executed collaboratively using multiple UAVs and involve various formations and scenario-based predefined advanced tasks, reducing the cognitive load for operators while retaining human supervision~\cite{4445761}.

\para{Formation} We consider the ability to dynamically switch formation patterns for different missions, as well as the ability to avoid collisions. Human operators can flexibly switch formations according to the environment and mission requirements, thus improving the adaptability of multi-UAVs. In our system, we provide three common formation options: 
\begin{inparaenum}
    \item V-shape, 
    \item Circle,
    \item Line.
\end{inparaenum}
The formation is shown in the top part of \figurename~\ref{fig:gesture-swarm-control}.

\para{Area Search} 
Our system enables area search operations where the human operator assigns search zones by providing target points and radius. During the mission, the operator can manually switch between predefined formations to adapt to different situations. This allows the user to focus on mission objectives rather than coordinating formations.

\para{Target Tracking} For reliable chasing of moving targets, we implement a coordinated tracking approach to enable adaptive trajectory alignment while avoiding inter-drone collisions. We assume the target is a UAV. Once the target enters this perception range, the tracking UAVs will start the synchronized tracking. We assume that this range is relatively large because, in actual operations, this might be implemented based on visual and other sensing methods.

\para{Space Coverage} For comprehensive coverage and data collection across extensive environmental regions, we employ a Voronoi diagram-based approach~\cite{1284411}. In this method, each UAV serves as a generating point for a Voronoi cell and relocates to the respective cell's location. This strategy not only facilitates the human operator in achieving efficient space coverage in open areas but also guarantees operational effectiveness.

\subsection{Embedded Algorithms and Controllers}
Our system integrates Reynolds' flocking~\cite{reynolds1987flocks, schilling2018learning} model for distributed formation control, leveraging the inherent separation and cohesion forces to enable collision avoidance and swarm cohesion. We enhance this foundation by combining velocity \textbf{alignment},\textbf{obstacle avoidance}, and \textbf{local repulsion mechanisms} to further advance collision prevention capabilities. To supplement the longer-range separation forces, we implement a half-spring model~\cite{vasarhelyi2018optimized} to generate short-range repulsive forces between nearby agents. The local repulsion acts as a secondary defense mechanism that works at shorter distances to avoid imminent collisions between dense, fast-moving agents.

\begin{figure*}[!t]
    \centering
    {\includegraphics[width=\linewidth]{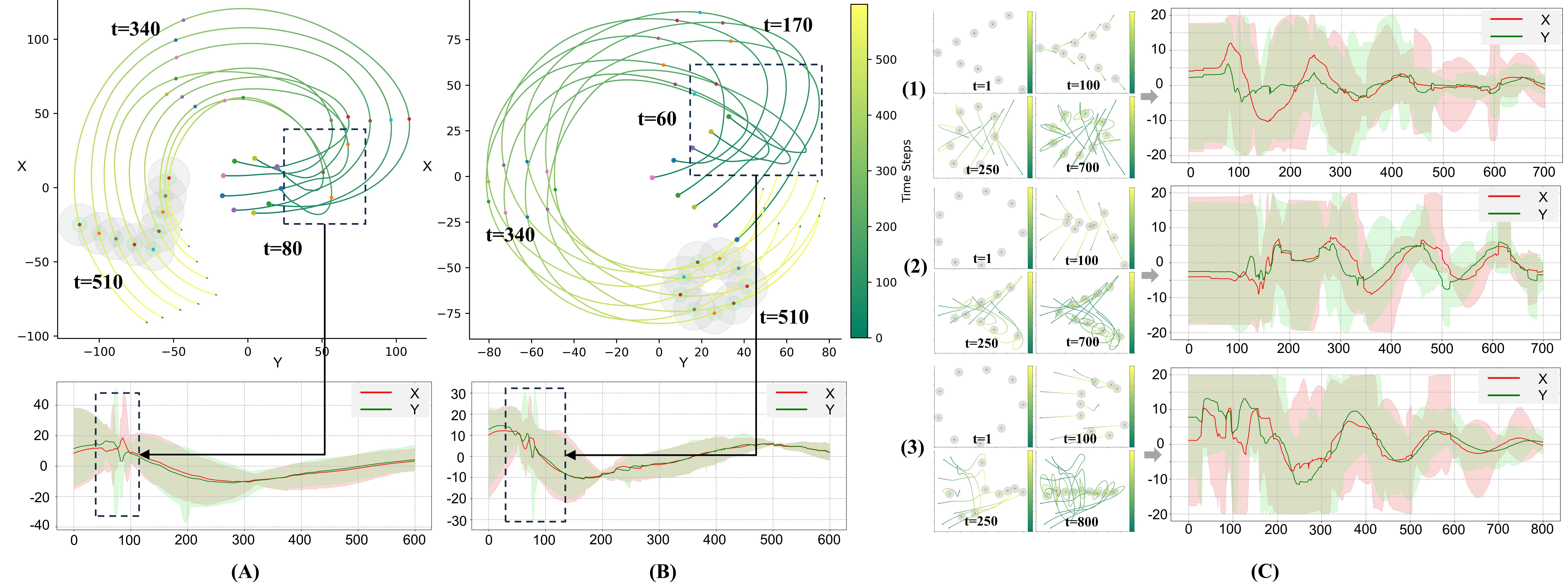}}
    \caption{
    Multi-UAV formation and coordination capabilities.
    \textbf{(A)} and \textbf{(B)}: formation maintenance and generation during the Search mission. UAV Trajectories over Time (Top) and Velocity Change over Time (Bottom); \textbf{(C)}: dynamic formation switching. (1) V to Circle, (2). Circle to V, and (3) Circle to the line. Trajectories over Time (Left) and Velocity change over time (Right). Note: The solid curve is the average of all UAV velocities, and the upper and lower distributions of the transparent part are the maximum and minimum values of velocities.}
    \label{fig:galaxy}
\end{figure*}

\section{Performance Evaluation}
\subsection{System Profiling Analysis}
\para{System Latency} 
We evaluated the latency implications of the Hand Gesture Recognition (HGR) system using a desktop computer. In real-time applications, latency can substantially degrade user experience and overall system efficiency. We measured response times across various modules, from initial hand detection to the final gesture decision. As detailed in \tablename~\ref{tbl:perf-profiling}, the results showcase time delays for each processing step. The most time-consuming module averaged $34\pm3$ ms, primarily attributed to extensive data processing from the hand's skeletal points. Conversely, the classification, conversion, and fusion modules consumed minimal time, ensuring a swift response after hand detection.

\para{System Accuracy} 
In our work, we adopted three separate algorithms for the classification of static and dynamic gestures. The unique signature of each gesture encompasses 16 static skeletal coordinates coupled with 2 dynamic skeletal sequences. Our dataset contains about 2,000 images for each gesture class, capturing variations in different gestures. As shown in \tablename~\ref{tab:model_eval}, the model shows good performance on the validation set. Both static and dynamic gestures achieved an accuracy rate of up to 99\%, which indicates that our model can recognize human gestures very accurately. 

\begin{table}[t!]
    \centering
    \begin{tabularx}{\columnwidth}{>{\centering\arraybackslash}X *{2}{>{\centering\arraybackslash}X}}
        \textbf{Hardware} & \textbf{Modules}&\textbf{Times(ms)}\\
        \hline
                                & Detect& 34 $\pm$ 3 \\
         GPU \textit{RTX 3080}  & \cellcolor{gray!25}Classify& \cellcolor{gray!25}$\leq$ 1 \\
       CPU \textit{Ryzen5 5600X}&  Convert & $\leq$ 1\\
                                &  \cellcolor{gray!25}Fusion  & \cellcolor{gray!25}$\leq$ 1 \\
        \hline
    \end{tabularx}
    \caption{We conducted a performance profiling analysis of the HGR system by running it 1,000 times and calculating the mean values. 
    }
    \label{tbl:perf-profiling}
\end{table}

\begin{table}[t!]
    \centering
    \begin{tabularx}{\columnwidth}{l l *{3}{>{\centering\arraybackslash}X}}
        \multicolumn{2}{c}{Static Classifier} & MLP & RNN & CNN \\
        \hline
        & Accuracy  & 0.98 & 0.99 & 0.99 \\
        \rowcolor{gray!25} &  F-1 Score & 0.99 & 0.99 & 0.99 \\
        & Precision & 0.99 & 0.99 & 0.99 \\
        \rowcolor{gray!25} &  Recall    & 0.99 & 0.99 & 0.99 \\
        \hline \\
        \multicolumn{2}{c}{Dynamic Classifier} & MLP & RNN & LSTM \\
        \hline
        & Accuracy  & 0.95 & 0.98 & 0.97 \\
        \rowcolor{gray!25} &  F-1 Score & 0.95 & 0.98 & 0.98 \\
        & Precision & 0.95 & 0.98 & 0.97 \\
        \rowcolor{gray!25} &  Recall    & 0.94 & 0.98 & 0.97 \\
        \hline
    \end{tabularx}
    \caption{Evaluation of static and dynamic classifier. For static gesture, we implemented models based on Multi-Layer Perceptron (MLP), Recurrent Neural Network (RNN), and Convolutional Neural Network (CNN). Conversely, for dynamic gesture classification, we select MLP, RNN, and Long Short-Term Memory (LSTM) networks.}
    \label{tab:model_eval}
\end{table}

\subsection{The Performance of mUAV Control System}
We measured our mUAV control system using the AirSim simulation environment~\cite{airsim2017fsr}, aiming to evaluate its capability in guiding the mUAV, particularly during complex behavioral shifts. We tracked the trajectory during formation transitions and observed changes in velocities at different steps during the flight. As demonstrated in \figurename~\ref{fig:galaxy}, the changes in velocity and trajectory indicate that each agent actively acts to prevent collisions and keep the formation. We then collected key metrics such as formation completion time, inter-collisions, velocity stability, and formation accuracy during the mission. The detailed results of these metrics are presented in \tablename~\ref{tab:formation_metrics}.

\subsection{Human User Study for System Performance Evaluation}
\textbf{Human User Study.} A 3-user pioneer study was conducted to improve questionnaire design; then 8 users were involved to assess HGIC system performance. HGIC was set up in our lab workstation to control a simulated multi-robot team; user assessment was conducted right after using the system; the questionnaire included questions about user experience on gesture and interface design (e.g., clarity of gesture classification,  convenience, and overall satisfaction); full questionnaire is here (link~\cite{userstudy2024}). The study results will be used for future system improvements.

\textbf{Result Analysis.} User study shows that participants gave positive feedback on various aspects of the developed HGIC system, detailed in Table III. Users agree on the clarity and logical reasonability of the gesture command design. Compared with traditional interaction methods using a keyboard and mouse, 87.5\% of people believed that gesture control is more convenient. 62.5\% of people said it was relatively easy to remember the gestures. 25\% of people are satisfied with the current version, while 75\% are satisfied with minor concerns on UI and function design, which will motivate us for continuous optimization. The overall evaluation indicated that users unanimously think the HGIC system as an interface will facilitate human-multi-robot interactions.

\begin{table}[]
\resizebox{\columnwidth}{!}{%
\begin{tabular}{cccccc}
 & Satisfied & Fairly Satisfied & Neutral & Less Satisfied & Dissatisfied \\ \hline
Clarity of Gestures & 62.5\% & 37.5\% & 0.0\% & 0.0\% & 0.0\% \\
\rowcolor{gray!25}Gesture Design Logic & 50.0\% & 50.0\% & 0.0\% & 0.0\% & 0.0\% \\
Motion Reasonableness & 62.5\% & 25.0\% & 12.5\% & 0.0\% & 0.0\% \\
\rowcolor{gray!25}\begin{tabular}[c]{@{}c@{}}Convenience Compared \\ with Traditional Way\end{tabular} & 37.5\% & 50.0\% & 12.5\% & 0.0\% & 0.0\% \\
Recognition Speed & 37.5\% & 25.0\% & 12.5\% & 25.0\% & 0.0\% \\
\rowcolor{gray!25} Memory Difficulty & 12.5\% & 62.5\% & 25.0\% & 0.0\% & 0.0\% \\
UI User-Friendliness & 50.0\% & 37.5\% & 12.5\% & 0.0\% & 0.0\% \\
\rowcolor{gray!25}Mode Switching Efficiency & 12.5\% & 75.0\% & 0.0\% & 0.0\% & 12.5\% \\
Overall Evaluation & 25.0\% & 75.0\% & 0.0\% & 0.0\% & 0.0\% \\ \hline
\end{tabular}%
}
    \caption{User satisfaction with the HGIC system. This table summarizes feedback on various attributes, such as clarity, logicality, difficulty in memorizing gestures, and mode-switching efficiency, indicating high user satisfaction with the entire system.}
    \label{User Study}
\end{table}

\section{Conclusion and Future Work}
In this paper, we developed an innovative system for mUAV control that utilizes camera-based gesture recognition, obviating the need for specialized hardware. Our method merges the intuitiveness of human gestures with the technical prowess required for managing mUAV. With the introduction of a mode-based gesture library combined with a multi-layered control architecture, our system guarantees that these commands are efficiently translated into complex behaviors across the mUAV, ensuring adaptability in ever-changing environments. For future works, the team will verify and evaluate the system in the field. 

\begin{table}[t]
\centering
\begin{tabularx}{\columnwidth}{l l *{3}{>{\centering\arraybackslash}X}}
 & Circle - V & V - Circle &  Circle - Line \\
\hline
Max Velocity (m/s) & 9.99 & 9.99 & 9.99 \\
\rowcolor{gray!25} 
Avg Velocity (m/s) & 4.55 & 5.94 & 6.94 \\
Duration (s) & 34.79 & 36.69 & 40.72 \\
\rowcolor{gray!25}Avg Spacing Error (m) & 1.17 & 1.68 & 1.26\\
Max Spacing Error (m) & 3.74 & 4.65 & 3.68\\
\rowcolor{gray!25} Collisions (times)  &2 & 0 & 0\\
\hline
\end{tabularx}
\caption{Evaluation of formation change tasks.}
\label{tab:formation_metrics}
\end{table}

\section{Acknowledgment}
We appreciate Han Chen for his insightful discussions. 

\addtolength{\textheight}{0cm}   

\bibliographystyle{ieeetr}
\bibliography{references}


\end{document}